\documentclass[environsciproc,article,submit,pdftex,moreauthors]{Definitions/mdpi} 

\preto{\abstractkeywords}{\nolinenumbers}
\newcommand{\urlNewWindow}[1]{\href[pdfnewwindow=true]{#1}{\nolinkurl{#1}}}

\usepackage{lscape}
\usepackage{environ}
\NewEnviron{myequation}{%
\begin{equation}
\scalebox{2}{$\BODY$}
\end{equation}
}

\firstpage{1} 
\pubvolume{1}
\issuenum{1}
\articlenumber{0}
\pubyear{2023}
\copyrightyear{2023}
\datereceived{ } 
\daterevised{ } 
\dateaccepted{ } 
\datepublished{ } 
\hreflink{https://doi.org/} 

\Title{Extending CAM-based XAI methods for Remote Sensing Imagery Segmentation}

\TitleCitation{Extending CAM-based XAI methods for Remote Sensing Imagery Segmentation}

\Author{Abdul Karim Gizzini$^{1}$, Mustafa Shukor $^{2}$ and Ali J. Ghandour\orcidA{} $^{3}$*}

\AuthorNames{Abdul Karim Gizzini, Mustafa Shukor and Ali J. Ghandou}

\AuthorCitation{Gizzini, A.; Shukor, M.; Ghandour, A.}

\address{%
$^{1}$ \quad Center for Digital Systems, IMT Nord Europe, Institut Mines-Télécom, University of Lille;\\
$^{2}$ \quad Sorbone University;\\
$^{3}$ \quad National Center for Remote Sensing - CNRS, Lebanon; aghandour@cnrs.edu.lb}

\corres{Corresponding author: Ali J. Ghandour, aghandour@cnrs.edu.lb;}

\abstract{
Artificial intelligence (AI) has recently covered many earth observation applications. AI-based data-driven methods perform exceptionally well in several remote sensing image processing tasks, such as object detection, classification, and segmentation.
However, current AI-based methods do not provide comprehensible physical interpretations of the used data, extracted features, and predictions/inference operations. As a result, deep learning models trained using high-resolution satellite imagery lack transparency and explainability and can be simply seen as a black box, which limits their wide-level adoption. Experts need help understanding the complex behavior of AI models and the underlying decision-making process. The field of explainable artificial intelligence (XAI) is an emerging field that provides means for a robust, practical, and trustworthy deployment of AI models.
Several XAI techniques have been proposed for image classification tasks, while the interpretation of image segmentation remains largely unexplored. This paper offers to bridge this gap by adapting the recent XAI classification algorithms and making them usable for muti-class image segmentation, where we mainly focus on buildings' segmentation from high-resolution satellite images.
To benchmark and compare the performance of the proposed approaches, we introduce a new XAI evaluation methodology and metric based on "Entropy" to measure the model uncertainty. Conventional XAI evaluation methods rely mainly on feeding area-of-interest regions from the image back to the pre-trained (utility) model and then calculating the average change in the probability of the target class. Those evaluation metrics lack the needed robustness, and we show that using Entropy to monitor the model uncertainty in segmenting the pixels within the target class is more suitable. 
We hope that this work will pave the way for additional XAI research for image segmentation and applications in the remote sensing discipline. Our code is publicly available \urlNewWindow{https://github.com/geoaigroup/GEOAI-ECRS2023}.}


\begin{document}

\section{Introduction} \label{introduction}

Recently, artificial intelligence (AI)-based models have been employed in a variety of computer vision tasks, from image classification, semantic segmentation, and object detection to image captioning, and visual question answering~\cite{zhang2023text}. These models mainly depend on convolutional neural networks (CNNs) that record superior performance. However, due to their complex deep architectures, CNNs are difficult to interpret~\cite{lipton2018mythos} and experts cannot understand the decision-making methodology of such models. 

Transparency is the ability of AI-based models to explain why they predict what they predict. Designing transparent models is crucial to build trust and pave the way for the integration of these systems into our daily lives. 
It was noted that better explainability and interpretability could be achieved using simple architectures with the cost of limited performance. In contrast,  using deep architectures sacrifices explainability to achieve better performance~\cite{jian2016deep}. 

Visual XAI methods are often used, where the aim is to highlight the parts of an input image that contributed the most to a particular prediction. Generally speaking, there are two main categories of XAI methods~\cite{nielsen2022robust}: (\textit{i}) perturbation-based or gradient-free methods, where the concept is to perturb input features (i.e., feature maps) by masking or altering their values, and record the effect of these changes on the model performance and (\textit{ii}) gradient-based methods where the gradients of the output (logits or soft-max probabilities) are calculated with respect to the extracted features or the input via back-propagation and used to estimate attribution scores. Gradient-based visual explanation methods are well known due to their computational efficiency. 

In classification problems, a good visual explanation of the model decision should localize the target class in the image (i.e., being class-specific), in addition to capturing fine-grained details within the target class. Therefore, the target class within the input image should be highlighted. However, this is not always the case in semantic segmentation problems, where spatial correlation between neighboring pixels within the input image should also be taken into consideration. \textbf{In other words, highlighting pixels within the target class may not be enough, since pixels outside the target class may also contribute to the model decision in the segmentation task. In this context, developing XAI methods for semantic segmentation is a challenging but not well-explored task.}

In this paper, we focus on buildings' rooftop segmentation models from high-resolution satellite images. A straightforward methodology is to adapt existing classification XAI methods toward semantic segmentation. The authors in~\cite{ref_seg_grad_cam} adapted the gradient-weighted class activation mapping (Grad-CAM) method~\cite{ref_grad_cam} that was originally proposed for classification to semantic segmentation. Inspired by the accomplished work in~\cite{ref_seg_grad_cam}, we adapt in this work a set of CAM-based XAI methods from classification to semantic segmentation. We also propose a new XAI evaluation metric that uses Entropy to measure the model uncertainty when feeding only the highlighted important regions, in addition to the target class pixels to the model.  To our knowledge, this work constitutes the first attempt to apply XAI gradient-based methods to remote sensing imagery segmentation. To sum up, the contribution of this paper is three-folds:

\begin{enumerate}
    \item Adapt five recently proposed CAM-based XAI methods from classification to semantic segmentation.
    \item Propose a new XAI evaluation methodology and metric that uses entropy to measure the model uncertainty.
    \item Benchmark the performance of the proposed XAI methods using the WHU dataset for buildings’ footprint segmentation from high-resolution satellite images.
\end{enumerate}

\section{Grad-CAM for Semantic Segmentation}
\label{gradcam_segmentation}

The convolutional layers retain spatial correlation, where the neurons capture the class-specific information in the image, i.e., parts related to the target object. In this context, the original Grad-CAM paper~\cite{ref_grad_cam} uses the gradient of the last convolutional layer to assign an importance indicator to each neuron for a particular desired decision. 

Let $\boldsymbol{A}^{k}$ be the $k^{th}$ feature map, where $1 \le k \le K$, and $K$ denote the total number of feature maps of the last convolutional layer of a classification network. Grad-CAM averages the gradients of the class of interest $c$ with respect to all $N$ pixels (indexed by $u, v$) of each feature map $\boldsymbol{A}^{k}$ to produce a weight $\alpha_{k}^{c}$ that denotes its importance as shown in Equation~\eqref{eq:grad_cam_weights}:

\begin{equation}
    \alpha_{k}^{c} = \frac{1}{N} \sum_{(u,v)} \frac{\partial y^{c}}{\partial \boldsymbol{A}_{u,v}^{k}},
    \label{eq:grad_cam_weights}
\end{equation}

where $y^{c}$ denotes the classification score of class $c$, and $\frac{\partial y^{c}}{\partial \boldsymbol{A}_{u,v}^{k}}$ denotes the gradients of $y^{c}$ with respect to $k-th$ feature maps. These gradients are calculated through backward propagation. After that, the feature maps are weighted with the calculated weights and passed to a ReLU function as shown in Equation~\eqref{eq:grad_cam_final}:

\begin{equation}
    \boldsymbol{L}_{\text{Grad-CAM}}^{c} = \text{ReLU} \Big(  \sum_{k} \alpha_{k}^{c} \boldsymbol{A}^{k} \Big).
\label{eq:grad_cam_final}
\end{equation}

Seg-Grad-CAM~\cite{ref_seg_grad_cam} modifies Equation ~\eqref{eq:grad_cam_weights} to be able to use it for the downstream segmentation task as shown in Equation~\eqref{eq:seg-grad-cam-weights}:

        \begin{equation}
            \alpha_{k}^{c} = \frac{1}{N} \sum_{(u,v)} \frac{\partial \sum_{(i,j) \in M} y_{i,j}^{c}}{\partial \boldsymbol{A}_{u,v}^{k}},
            \label{eq:seg-grad-cam-weights}
        \end{equation}

        where $M$ is the set of pixel indices ($i,j$) of interest in the output segmented mask. After calculating the weights, Seg-Grad-CAM proceeds according to Equation~\eqref{eq:grad_cam_final}. Hence, Seg-Grad-CAM highlights the important pixels that contribute to the segmentation decision of the considered region of interest.
\section{CAM-based Extensions} \label{extensions}

This section sheds light on the five recent CAM-based extensions, showing their main limitations and corresponding key enhancements. We adapted all these five methods from classification to semantic segmentation following the same approach shown in~\eqref{eq:seg-grad-cam-weights}, that is, replacing the classification score with the segmentation scores of the target class.

\begin{enumerate}
    \item Seg-Grad-CAM++~\cite{gradcampp}: has been designed to address the limitation of Grad-CAM that lies in localizing multiple occurrences of the same object (class) within the input image. This could be addressed by taking a weighted average of the pixel-wise gradients, where Equation~\eqref{eq:grad_cam_weights} can be rewritten as Equation~\eqref{eq:grad_campp_weights}:

            \begin{equation}
                \alpha_{k}^{c} =  \sum_{(u,v)} w_{u,v}^{k,c} ~\text{ReLU}\big(\frac{\partial y^{c}}{\partial \boldsymbol{A}_{u,v}^{k}}\big),
                \label{eq:grad_campp_weights}
            \end{equation}

            where $w_{u,v}^{k,c}$ are the weighting coefficients of the pixel-wise gradients for class $c$ and feature map $\boldsymbol{A}^{k}$. Therefore, $w^{k,c}$ captures the importance of a particular activation map, ensuring that all the features maps related to the target class are highlighted with equal importance. 

            \item Seg-XGrad-CAM: which adapts Axiom-based Grad-CAM~\cite{fu2020axiom} to the segmentation realm. The main enhancement in~\cite{fu2020axiom} is in the calculation of the importance weight of the feature map by solving an optimization problem that meets the sensitivity and conservation constraints. The optimal $\alpha_{k}^{c}$ is expressed in Equation~\eqref{eq:xgradcam}:
            \begin{equation}
               \alpha_{k}^{c} = \sum_{(u,v)} \Big( \frac{\boldsymbol{A}_{u,v}^{k}}{\sum_{(u,v)} \boldsymbol{A}_{u,v}^{k}}  \frac{\partial y^{\prime c}}{\partial \boldsymbol{A}_{u,v}^{k}}\Big)
                \label{eq:xgradcam}
            \end{equation}

            where $y^{\prime c}$ denotes the sum of the element-wise product of feature maps and the gradient maps of the target layer. 

            \item Seg-Score-CAM: which adapts Score-CAM~\cite{scorecam} from classification to segmentation. Gradients are noisy and may not be an optimal solution for highlighting important regions within the input image. Hence, the "Increase in Confidence" criteria is used in~\cite{scorecam} to quantify the feature map importance, where the feature map weight $\alpha_{k}^{c}$ is calculated in Equation~\eqref{eq:score_cam_final}:
            \begin{equation}
                    \alpha_{k}^{c} = C(\boldsymbol{A}^{k}).
            \label{eq:score_cam_final}
            \end{equation} 

            where $C(.)$ denotes the increase in confidence score for the feature map considered $\boldsymbol{A}^{k}$. This can be calculated by perturbing the input image by $\boldsymbol{A}^{k}$, where the importance of this activation map is obtained by the target score of the perturbated input image. 
          
        \item Seg-Ablation-CAM: which adapts Ablation-CAM~\cite{Ablation_CAM} to segmentation. Backpropagated gradients do not retain spatial information related to the target class. Thus, the gradient-free Ablation-CAM method~\cite{Ablation_CAM} is based on the idea of calculating $y_{k}^{c}$ in the presence of the feature map $\boldsymbol{A}^{k}$ and repeating the forward pass of the original input image but with zero feature maps. Therefore, the output score $y^{c}$ is reduced compared to $y_{k}^{c}$ and acts as a baseline. In this context, the "importance weight" of each feature map can be computed as shown in Equation~\eqref{eq:ablcam1}:

            \begin{equation}
                \alpha_{k}^{c} = \frac{y^{c} - y_k^{c}}{y^{c}}.
                \label{eq:ablcam1}
            \end{equation}

         The "importance weight" can be interpreted as the drop in the class $c$ score when the feature map $\boldsymbol{A}^{k}$ is removed. 

        \item Seg-Eigen-CAM: which finally adapts Eigen-CAM~\cite{eigen_cam} to the task at hand. Work in~\cite{eigen_cam} is based on the principal components of the learned representations from the convolutional layers without the need for backpropagation. Let $\boldsymbol{O}$ be the output of the semantic segmentation of the target class, where its principal components can be computed by factorizing using the singular value decomposition. $\boldsymbol{L}_{\text{Seg-Eigen-CAM}}^{c}$ is then computed as follows in Equation~\eqref{eq:eigen_cam_final}:  

        \begin{equation}
        \boldsymbol{L}_{\text{Seg-Eigen-CAM}}^{c} = \boldsymbol{O} \boldsymbol{V}_{1},~~~~~ \boldsymbol{O} = \boldsymbol{U} \boldsymbol{\Sigma} \boldsymbol{V}^{T}.
        \label{eq:eigen_cam_final}
        \end{equation}

        $\boldsymbol{U}$ represents the left singular vector, $\boldsymbol{V}$ refers to the right singular vector with $V_{1}$ as the first eigenvector, and $\boldsymbol{\Sigma}$ is a diagonal matrix with singular values along the diagonal.

\end{enumerate} 
\section{XAI Evaluation} 

This section sheds light on the performance evaluation of the studied XAI methods. First, we introduce the proposed theoretical evaluation framework, where we define the evaluation methodologies used in addition to the proposed entropy evaluation metric and discuss the results.

\label{results}
\subsection{Proposed Framework}

\begin{figure*}[t]
\centering
\includegraphics[width=\textwidth,height=\textheight,keepaspectratio]{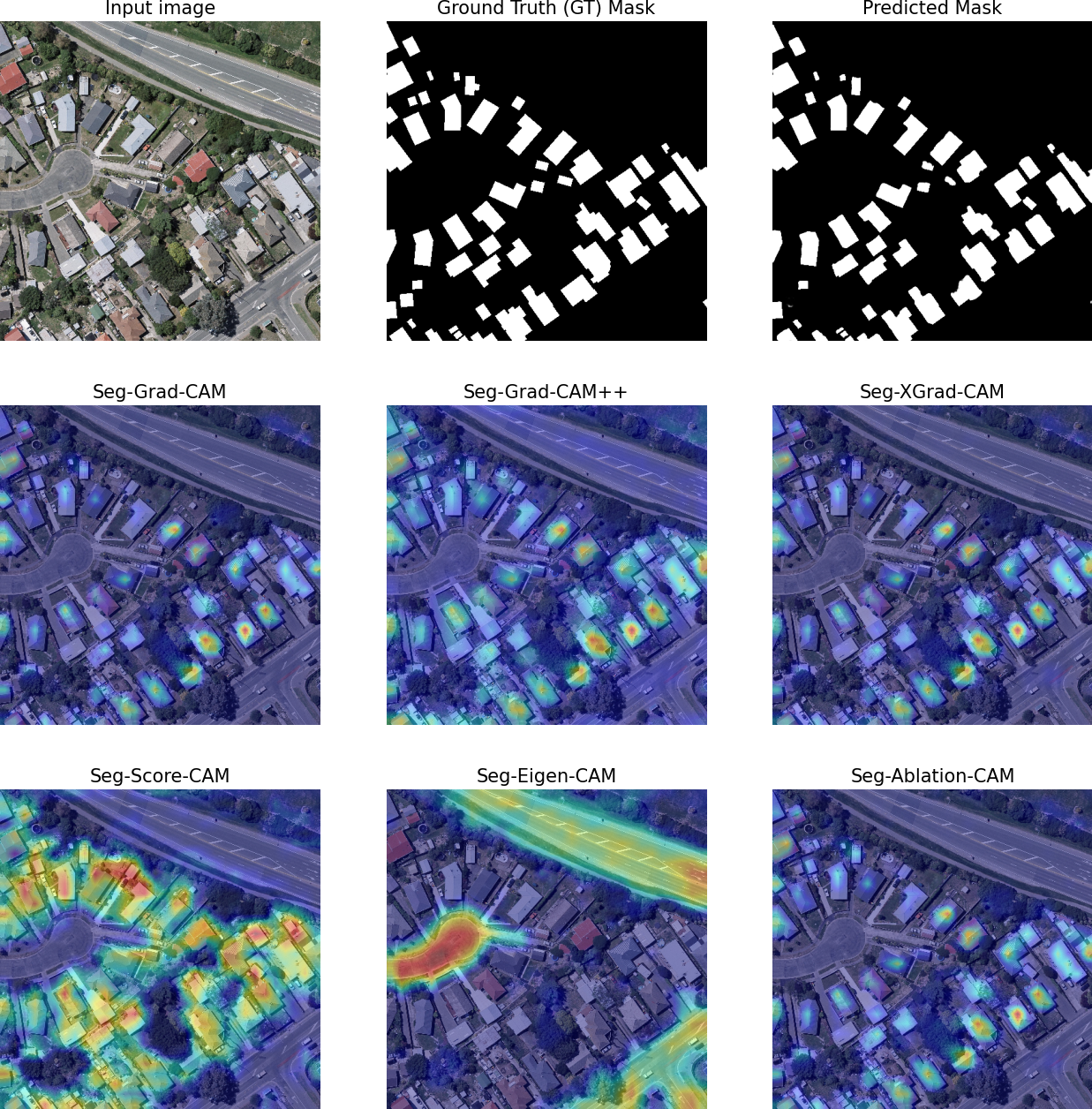}
    \caption{Saliency map of the studied CAM-based XAI methods for semantic segmentation. The target class is "buildings" and the inspected layer is the first UNet decoder block.}
    \label{fig:enter-label}
\end{figure*}

Performance evaluation of XAI methods, including the methodology and evaluation metric employed, has been an active research topic. In classification problems, XAI evaluation is based on two main methodologies:

\begin{itemize}
    \item M1 - Background only: where the highlighted relevant pixels are masked from the original image, and then the masked image is fed again to the pre-trained model, where the drop in the classification score is expected. A higher drop in the classification score signifies that the corresponding XAI method is better, as the model cannot preserve the previously achieved classification score when the relevant pixels are masked.

    \item M2 - Highlighted only: In contrast to M1, here the idea is to mask the background and keep only the relevant pixels highlighted. In this case, an increase in confidence is expected due to the fact that we are feeding the model with the pixels needed only to perform the classification task. We note that in some cases there is no increase in the classification score due to masking of a large part of the input image. In this case, a lower drop in the classification score means that the corresponding XAI method is better.

It is worth mentioning that monitoring the change in the segmentation score employing M1 and M2 could be applied in the semantic segmentation task, but this is not enough due to the spatial correlation between neighboring pixels. In other words, masking pixels from the input image may impact the model's decision to segment other classes.
\textbf{XAI method for the segmentation task is expected to highlight relevant pixels, regardless of whether these relevant pixels are part of the target class or any other classes. Therefore, monitoring the change in the segmentation score of the target class is not enough to measure the performance of an XAI method.}

To handle this challenge, we suggest the following evaluation methodology denoted as M3 for semantic segmentation.\\

    \item M3 - Highlighted + Target:
Let $\boldsymbol{T} \subseteq \boldsymbol{\Omega}$ be the set of pixels belonging to the target class $c$, where $\boldsymbol{\Omega}$ refers to the universe (entire set) of pixels of a given input image $\boldsymbol{I}$. Let $\boldsymbol{\Phi}$ be the set of relevant pixels highlighted by the XAI method.  
The perturbated image $\boldsymbol{I}^{\prime}$ to be passed to the model is defined as $\boldsymbol{I}^{\prime} = \Phi \cup \boldsymbol{T}$.
\end{itemize}

Based on M3, we propose the use of Shannon's entropy~\cite{shannon1948mathematical} as an evaluation metric of XAI. Given an input image $\boldsymbol{I}$, the pixel-wise entropy map $\boldsymbol{E}_{\boldsymbol{I}}$ can be calculated as depicted in Equation~\eqref{eq:entropy}:

\begin{equation}
    E_{\boldsymbol{I}}^{(i,j)} = -\frac{1}{\log(L)} \sum_{l = 1}^{L}  P_{\boldsymbol{I}}^{(i,j,l)} log\Big(P_{\boldsymbol{I}}^{(i,j,l)}\Big).
    \label{eq:entropy}
\end{equation}

$P_{\boldsymbol{I}}$ denotes the $L$-diemnsional softmax output of the model under consideration. The propposed entropy metric $E_{\text{XAI}}$ is calculated as shown in Equation~\eqref{eq:entropy_XAI}:

\begin{equation}
    E_{\text{XAI}} = \sum_{i,j} E_{\boldsymbol{I}}^{(i,j)}.
    \label{eq:entropy_XAI}
\end{equation}

We note that $E_{\text{XAI}}$ measures the model's uncertainty in segmenting the target class's pixels. We note that a lower increase in the $E_{\text{XAI}}$ means that the pre-trained model can better classify the target pixels within the target class when it sees (highlighted pixels + target class pixels) only.

\begin{table*}[t]
\renewcommand{\arraystretch}{1.3}
\centering
\caption{Evaluation of XAI methods for semantic segmentation using three Metrics/Methodologies pairs: (\textit{i}) Drop in Segmentation Score/M1 - Background only (higher is better), (\textit{ii}) Drop in Segmentation Score/M2 - Highlighted only (lower is better), and (\textit{iii}) increase in entropy/M3 - Highlighted + Target (lower is better).} We note that the reported results are for the WHU test set. The average Segmentation Score (SS) and entropy are $89.11\%$ and $0.000898$, respectively, for the entire input image $\boldsymbol{I}$.
\label{tb:exec_time_soa}
\resizebox{\textwidth}{!}{%
\begin{tabular}{|c|c|c|c|c|c|c|}
\hline
\textbf{XAI Method}                                                               & \textbf{Seg-Grad-CAM} & \textbf{Seg-Grad-CAM++} & \textbf{Seg-XGrad-CAM} & \textbf{Seg-Score-CAM} & \textbf{Seg-Eigen-CAM} & \textbf{Seg-Albation-CAM} \\ \hline
\begin{tabular}[c]{@{}c@{}}(\%) Drop in SS/M1\\ (higher is better)\end{tabular}   & 4.80\%                & 10.32\%                 & 5.04\%                 & \textbf{13.04\%}       & 9.19\%                 & 5.23\%                    \\ \hline
\begin{tabular}[c]{@{}c@{}}(\%) Drop in SS/M2\\ (lower is better)\end{tabular}    & 56.51\%               & 49.76\%                 & 56\%                   & \textbf{45.96\%}       & 47.49\%                & 55.51\%                   \\ \hline
\begin{tabular}[c]{@{}c@{}}(\%) Increase in $E_{\text{XAI}}$/M3\\ (lower is better)\end{tabular} & 43.65\%               & 58.12\%                 & 43.82\%                & \textbf{33.63\%}       & 71.49\%                & 43.65\%                   \\ \hline
\end{tabular}
}
\end{table*}

\subsection{Results}

In this subsection, we rely on the rooftop segmentation model~\cite{model} trained using the WHU dataset. The baseline segmentation score and entropy correspond to the segmentation results when the full input image $\boldsymbol{I}$ is fed to the model considered. Table 1 shows results for studied M1 (Backgorund only), M2 (Highlighted only) XAI evaluation methodologies where the drop in the Segmentation Score (SS) is recorded. In addition to the proposed M3 (Highlighted + Target class) XAI evaluation methodology that employs $E_{\text{XAI}}$ as an evaluation metric.

As mentioned earlier, the higher the drop when using M1, the better. The lower the drop when employing M2, the better. We can observe in row 2 of Table 1 a large drop in the segmentation score for all six XAI methods when using M2 - Highlighted only, which comes as a surprise and is attributed to the fact that a large part of the input image is being masked (after thresholding).

When employing M1, Seg-Score-CAM is the best performer and records the highest drop in segmentation score. Seg-XGrad-CAM is the worst performer and records the lowest drop in the segmentation score. When using M2, Seg-Score-CAM again is the best performer and records the lowest drop in segmentation score. Seg-Grad-CAM is the worst performer and records the highest drop in the segmentation score.

On the other hand, when using M3, Seg-Score-CAM is the best performer and records the lowest increase in entropy. Seg-Eigen-CAM records the highest increase in entropy, which means that it has the worst performance among the XAI methods studied. Therefore, the pixels highlighted by Seg-Eigen-CAM are not the important ones used by the model to segment rooftops according to Entropy/M3.

Therefore, in summary, the results of M3 in Table 1 confirm the findings of M1 and M2 that Seg-Score-CAM returns the best explanation. The qualitative results of Seg-Score-CAM in Figure 1 reveal that the most relevant pixels are located towards the buildings' boundaries. This finding is directly related to the observation stated earlier: In the segmentation task, pixels outside the target class might contribute to the model decision.
Furthermore, the tabulated results for M1, M2 and M3 are consistent in that Seg-Grad-CAM, Seg-XGrad-CAM, and Seg-Abblation-CAM show comparable performance.

The main difference is related to Seg-Eigen-CAM, which is found to perform poorly (the worst) under entropy and M3, while it is amongst the best performers using M1 and M2. The qualitative results in Figure 1 confirm that the Seg-Eigen-CAM saliency map did not do a good job explaining the behavior of the underlying model. This case study clearly shows the importance of the proposed entropy metric and the M3 evaluation methodology when evaluating segmentation-based XAI methods.
\section{Conclusions}

This paper sheds light on the adaptation of recent gradient-based and gradient-free XAI methods for semantic segmentation tasks, with a particular focus on buildings’ segmentation from high-resolution satellite images. We proposed a novel XAI evaluation methodology and metric based on entropy to measure the model uncertainty in segmenting the target class pixels. The results show that the gradient-free Seg-Score-CAM method outperforms the other benchmarked methods. As a future perspective, we will investigate the ability to design hybrid XAI methods, in addition to extending the XAI evaluation methodologies to cover more reliable interpretations. 
\vspace{6pt} 



\authorcontributions{

Conceptualization, Abdul Karim Gizzini, Mustafa Shukor, and Ali J. Ghandour; Data curation, Abdul Karim Gizzini; Formal analysis, Abdul Karim Gizzini, Mustafa Shukor, and Ali J. Ghandour; Investigation, Abdul Karim Gizzini; Methodology, Abdul Karim Gizzini, and Mustafa Shukor; Project administration, Ali J. Ghandour; Resources, Ali J. Ghandour; Software, Abdul Karim Gizzini; Supervision, Ali J. Ghandour; Validation Abdul Karim Gizzini, Mustafa Shukor, and Ali J. Ghandour; Visualization, Abdul Karim Gizzini; Writing - original draft, Abdul Karim Gizzini; Writing - review \& editing,  Mustafa Shukor, and Ali J. Ghandour;



}

\funding{``This research received no external funding''.}

\conflictsofinterest{``The authors declare no conflict of interest.''}

\begin{adjustwidth}{-\extralength}{0cm}

\reftitle{References}


\bibliography{sample}

\PublishersNote{}
\end{adjustwidth}
\end{document}